\documentclass[runningheads]{llncs}

\usepackage{graphicx}
\usepackage{verbatim}
\usepackage{amsfonts}
\usepackage{xcolor}
\usepackage{float}
\usepackage{algorithmic}
\usepackage{algorithm}
\usepackage{amsmath}
\usepackage{booktabs}
\DeclareMathOperator*{\argmax}{argmax}
\usepackage{tabulary}
\usepackage{colortbl}
\usepackage{etoolbox}
\usepackage{multicol}
\usepackage{hyperref}
\usepackage{lipsum}
\usepackage{subcaption}
\usepackage{wrapfig}
\usepackage{makecell}
\usepackage{capt-of}
\usepackage{makecell}

\usepackage[T1]{fontenc}
\usepackage[utf8]{inputenc}
\usepackage[font=small,labelfont=bf]{caption}

\begin{document}
\title{Globally-Aware Multiple Instance Classifier for Breast Cancer Screening}
%
%

\author{Yiqiu Shen\inst{1} \and
Nan Wu\inst{1} \and
Jason Phang\inst{1} \and
Jungkyu Park\inst{1} \and
Gene Kim\inst{2} \and\\
Linda Moy\inst{2} \and
Kyunghyun Cho\inst{1,3,4,5} \and
Krzysztof J. Geras\inst{2,1}
}
\authorrunning{Y. Shen et al.}
%
\institute{Center for Data Science, New York University \and
Department of Radiology, New York University School of Medicine \and
Department of Computer Science, Courant Institute, New York University \and
Facebook AI Research \and
CIFAR Azrieli Global Scholar}

\maketitle              
\begin{abstract}
Deep learning models designed for visual classification tasks on natural images have become prevalent in medical image analysis. However, medical images differ from typical natural images in many ways, such as significantly higher resolutions and smaller regions of interest. Moreover, both the global structure and local details play important roles in medical image analysis tasks. To address these unique properties of medical images, we propose a neural network that is able to classify breast cancer lesions utilizing information from both a global saliency map and multiple local patches. The proposed model outperforms the ResNet-based baseline and achieves radiologist-level performance in the interpretation of screening mammography. Although our model is trained only with image-level labels, it is able to generate pixel-level saliency maps that provide localization of possible malignant findings.

\keywords{deep learning \and neural networks \and breast cancer screening \and weakly supervised localization \and high-resolution image classification}
\end{abstract}

\section{Introduction}
As the second leading cause of cancer death among women in the US, breast cancer has been studied for decades. While studies have shown screening mammography has significantly reduced breast cancer mortality, it is an imperfect tool \cite{RN40}. To address its limitations, convolutional neural networks (CNN) designed for computer vision tasks on natural images have been applied. For instance, VGGNet \cite{simonyan2014very}, designed for object classification on ImageNet \cite{imagenet_cvpr09}, has been applied to breast density classification \cite{wu2017breast} and Faster R-CNN \cite{ren2015faster} has been adapted to localize suspicious findings in mammograms \cite{ribli2018detecting}. We refer the readers to \cite{gao2019new} for a comprehensive review of prior work on machine learning for mammography.

The compatibility between the models designed for natural images and the distinct properties of medical images remains an open question. Firstly, medical images are usually of a much higher resolution than typical natural images, so deep CNNs that work well for natural images may not be applicable to medical images due to GPU memory constraints. Moreover, for many applications, regions of interest (ROI) in medical images, such as lesions and calcifications, are proportionally smaller in size compared to those in natural images. Fine details, often only a few pixels in size, along with global features such as the spatial distribution of radiodense tissue determine the labels. In addition, while natural images can be aggressively downsampled and preserve the information necessary for classification, significant amounts of information could be lost from downsampling medical images, making the correct diagnosis unattainable.

\textbf{Contributions.} \enspace In this work, we address the aforementioned issues by proposing a novel model for the classification of medical images. The proposed model preserves global information in a saliency map and aggregates important details with a Multiple Instance Learning (MIL) framework. Unlike existing approaches that rely on pixel-level lesion annotations \cite{wu2019deep,ribli2018detecting}, our model only requires image-level supervision and is able to generate pixel-level saliency maps that highlight suspicious lesions. In addition, our model is equipped with an attention mechanism that enables it to select informative image patches, making the classification process interpretable. When trained and evaluated on more than 1 million high-resolution breast cancer screening exams, our model outperforms a ResNet-based baseline \cite{wu2019deep} and achieves radiologist-level performance.

\textbf{Related Works.} \enspace 
Existing methods have approached the breast cancer detection problem using techniques such as MIL \cite{zhu2017deep} and 3D CNNs \cite{wang2018densely}. Our model is inspired by works on weakly supervised object detection. Recent progress demonstrates that CNN classifiers, trained with image-level labels, are able to perform semantic segmentation at the pixel level \cite{diba2017weakly,durand2017wildcat,yao2018weakly}. This is achieved in two steps. First, a backbone CNN converts the input image to a saliency map (SM) which highlights the discriminative regions. A global pooling operator then collapses the SM into scalar predictions which makes the entire model trainable end-to-end. To make an image-level prediction, most existing models rely on the SM which often neglects fine-grained details. In contrast, our model also leverages local information from ROI proposals using a dedicated patch-level classifier. In Section \ref{cp}, we demonstrate that the ability to focus on fine visual detail is important for classification. 

\begin{figure}
  \centering
 \includegraphics[width=1.1\textwidth]{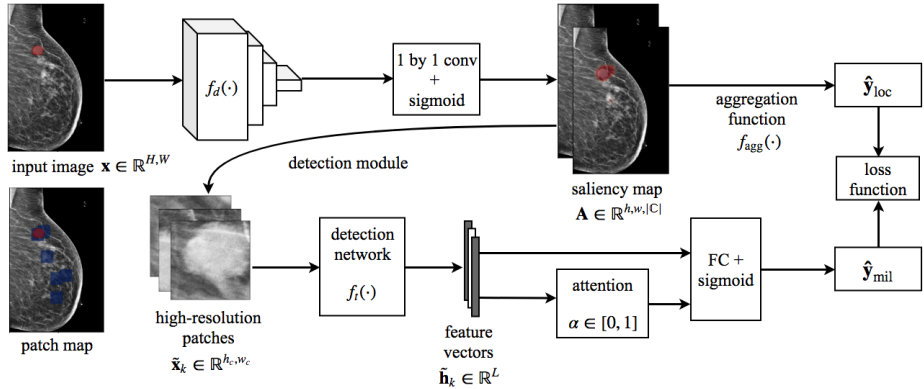}
 \vspace{-3mm}
  \caption{Overall architecture of GMIC. The input image is annotated with true ROIs (red). The patch map indicates positions of ROI patches (blue squares) on the input.}
  \label{overall_plot}
\end{figure}

\section{Methods}
We formulate our task as a multi-label classification. Given a grayscale high-resolution image $\mathbf{x} \in \mathbb{R}^{H,W}$, we would like to predict the label $\mathbf{y}$, where $y^c$ denotes whether class $c \in \mathbb{C}$ is present. As shown in Figure \ref{overall_plot}, the Globally-Aware Multiple Instance Classifier (GMIC) consists of three modules: (i) The localization module processes $\mathbf{x}$ to generate a SM, denoted by $\mathbf{A}$, which indicates approximate localizations of ROIs. (ii) The detection module uses $\mathbf{A}$ to retrieve $K$ patches from $\mathbf{x}$ as refined proposals for ROIs. (iii) We use an MIL framework to aggregate information from retrieved patches and generate the final prediction.

\subsection{Localization Module} \label{lmod}
As illustrated in Figure \ref{overall_plot}, the localization module first uses a CNN $f_d(\cdot)$ to extract relevant features from $\mathbf{x}$. Due to memory constraints, input images are usually down-sampled before $f_d(\cdot)$ \cite{yao2018weakly}. For mammograms, however, down-sampling distorts important visual details such as lesion margins and blurs small ROIs. In order to retain the original resolution, we parameterize $f_d(\cdot)$ as a ResNet-22 \cite{wu2019deep} and remove its global average pooling and fully connected layers. This model has fewer filters than the original ResNet architectures in each layer in order to process the image at the full resolution while keeping GPU memory consumption manageable. The feature maps obtained after the last residual block are transformed into the SM $\mathbf{A} \in \mathbb{R}^{h,w,|\mathbb{C}|} $ using $1\times1$ convolution with sigmoid non-linearity. Each element of $\mathbf{A}$,  $\mathbf{A}_{i,j}^c \in [0,1]$, denotes a score that indicates the contribution of spatial location $(i,j)$ towards classifying the input as class $c$.

\subsection{Detection Module} \label{dmod} 
Due to its limited width, $f_d(\cdot)$ is only able to provide coarse localization. We propose using patches as ROI proposals to complement the localization module with fine-grained detail. We designed a greedy algorithm (Algorithm \ref{alg:roi}) to retrieve $K$ proposals for ROIs, $\tilde{\mathbf{x}}_k \in \mathbb{R}^{h_c,w_c}$, from the input $\mathbf{x}$. In our experiments, we set $K = 6$, and $w_c = h_c = 256$. The reset rule in line 12 explicitly ensures that extracted ROI proposals do not significantly overlap with each other.

\begin{algorithm*}[t]
    \caption{Retrieve the ROIs}
    \label{alg:roi}
    \vspace{-13pt}
    \begin{multicols}{2}
    \begin{algorithmic}[1]
        \REQUIRE  $\mathbf{x} \in \mathbb{R}^{H,W}$, $\mathbf{A} \in \mathbb{R}^{h,w,|\mathbb{C}|}$, $K$
        \ENSURE $ O = \{ \tilde{\mathbf{x}}_k |  \tilde{\mathbf{x}}_k \in \mathbb{R}^{h_c,w_c} \}$
        \STATE{$O = \emptyset$}
        \FOR{each class $c \in \mathbb{C}$}
            \STATE{$\mathbf{\tilde{A}}^c = \text{min-max-normalization}(\mathbf{A}^c)$}
         \ENDFOR\\
         \STATE{$ \hat{\mathbf{A}} = \sum_{c \in \mathbb{C}} \tilde{\mathbf{A}}^c$}
         \STATE{$l$ denotes an arbitrary $h_c \frac{h}{H} \times w_c \frac{w}{W}$ rectangular patch on $\mathbf{\hat{A}}$}
         \STATE $f_c(l, \hat{\mathbf{A}}) = \sum_{(i,j) \in l} \hat{\mathbf{A}}[i,j]$
        \FOR{each $1,2,...,K$}
           
            \STATE{$l^* = \argmax_{l} f_c(l, \hat{\mathbf{A}})$}
            \STATE{$L = $ position of $l^*$ in $\mathbf{x}$}
            \STATE{$O = O \cup \{L\}$}
            \STATE{$\hat{\mathbf{A}}[i,j]=0, \forall (i,j) \in l^*$}
        \ENDFOR
        \RETURN $O$
    \end{algorithmic}
  \end{multicols}
  \vspace{-8pt}
\end{algorithm*}

\subsection{Multiple Instance Learning Module} \label{mil}
Since ROI patches are retrieved using a coarse saliency map, the information relevant for classification carried in each patch varies significantly. To address this, we apply an MIL framework to aggregate information from ROI patches. A detection network $f_t(\cdot)$ is first applied on every instance $\tilde{\mathbf{x}}_k$ and converts them into feature vectors $\tilde{\mathbf{h}}_k \in \mathbb{R}^L$. We use $L = 128$ in all experiments. We parameterize $f_t(\cdot)$ as a ResNet-18 (pretrained on ImageNet \cite{imagenet_cvpr09}). Since not all ROI patches are relevant to the prediction, we use the Gated Attention Mechanism proposed in \cite{ilse2018attention} to let the model select informative patches. The selection process yields an attention-weighted representation $\mathbf{z} = \sum_{k=1}^{K} \alpha_k \tilde{\mathbf{h}}_k$, where attention score $\alpha_k \in [0,1]$ indicates the relevance of each patch $\tilde{\mathbf{x}}_k$. The representation $\mathbf{z}$ is then passed to a fully connected layer with sigmoid activation to generate a prediction $\hat{\mathbf{y}}_{\text{mil}} = \text{sigm}(\mathbf{w_{\text{mil}}}^T \mathbf{z})$, where $ \mathbf{w}_{\text{mil}} \in \mathbb{R}^{L \times |\mathbb{C}|}$ are learnable parameters.

\subsection{Training} \label{td}
It is difficult to make this model trainable end-to-end. Since the detection module is not differentiable, the gradient from the training loss $L(\mathbf{y},\hat{\mathbf{y}}_\text{mil})$ will not flow into the localization module. Inspired by \cite{diba2017weakly}, we circumvent this problem with a scheme that simultaneously trains the localization module and the MIL module. An aggregation function $f_\text{agg}(\mathbf{A}^c): \mathbb{R}^{h,w} \mapsto [0,1]$ is designed to map the SM for each class $c$ into a prediction $\hat{\mathbf{y}}_\text{loc}^c$. The design of $f_\text{agg}(\mathbf{A}^c)$ has been extensively studied \cite{durand2017wildcat}. Global Average Pooling (GAP) would dilute the prediction as most of the spatial locations in $\mathbf{A}^c$ correspond to background and provide little training signal. On the other hand, Global Max Pooling (GMP) only backpropagates gradient into a single spatial location which makes the learning process slow and unstable. In our work, we use a soft balance between GAP and GMP : $f_{\text{agg}}(\mathbf{A}^c) = \frac{1}{|H^+|}\sum_{(i,j) \in H^+} \mathbf{A}^c_{i,j}$, where $H^+$ denotes the set containing locations of top $t\%$ values in $\mathbf{A}^c$, and $t$ is a hyper-parameter. The prediction $\hat{\mathbf{y}}_\text{loc}^c = f_{\text{agg}}(\mathbf{A}^c)$ is a valid probability as $\mathbf{A}^c_{i,j} \in [0,1]$. To fine-tune the SM and prevent the localization module from highlighting irrelevant areas, we impose the following regularization on $\mathbf{A}^c$: $L_\text{reg}(\mathbf{A}^c) = \sum_{(i,j)} |\mathbf{A}^c_{i,j}|^{\beta}$, where $\beta$ is a hyper-parameter. In summary, the loss function used to train the entire model is:
\begin{equation}
    L(\mathbf{y}, \hat{\mathbf{y}}) = \sum_{c \in \mathbb{C}} \text{BCE}(\mathbf{y}^c, \hat{\mathbf{y}}_{\text{loc}}^c) + \text{BCE}(\mathbf{y}^c, \hat{\mathbf{y}}_{\text{mil}}^c) + \lambda L_\text{reg}(\mathbf{A}^c),
\end{equation}
where $\text{BCE}(\cdot , \cdot)$ is the binary cross-entropy and $\lambda$ is a hyper-parameter. In the inference stage, the prediction is computed as $\hat{\mathbf{y}} = \frac{1}{2} (\hat{\mathbf{y}}_{\text{mil}} + \hat{\mathbf{y}}_{\text{loc}})$.

\section{Experiments}
\begin{wrapfigure}{r}{0.30\textwidth}
\vspace{-23pt}
        \begin{tabular}{c c}
  \reflectbox{\includegraphics[width=0.14\textwidth]{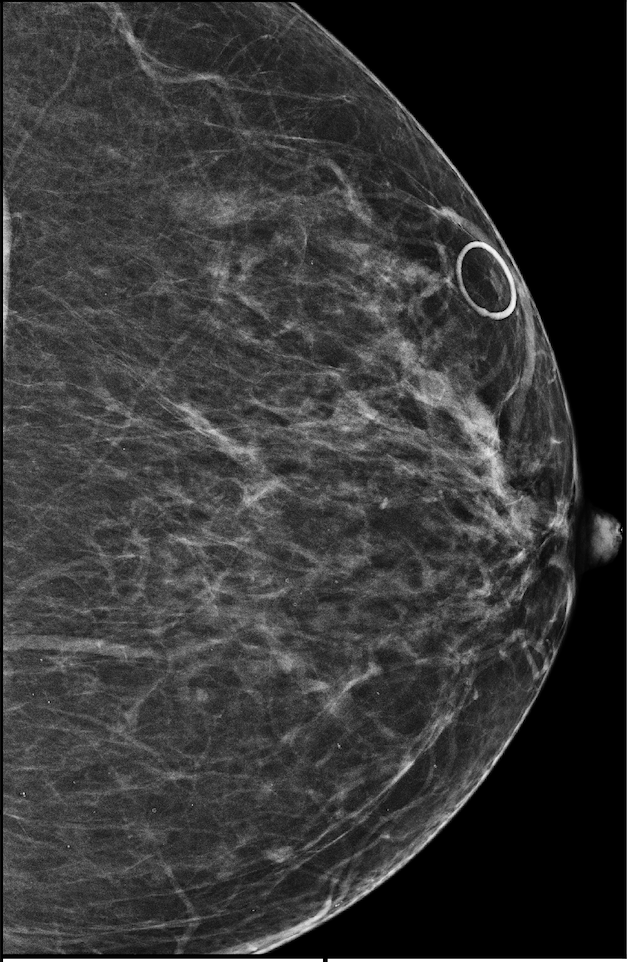}}\vspace{-0.75mm} & \includegraphics[width=0.14\textwidth]{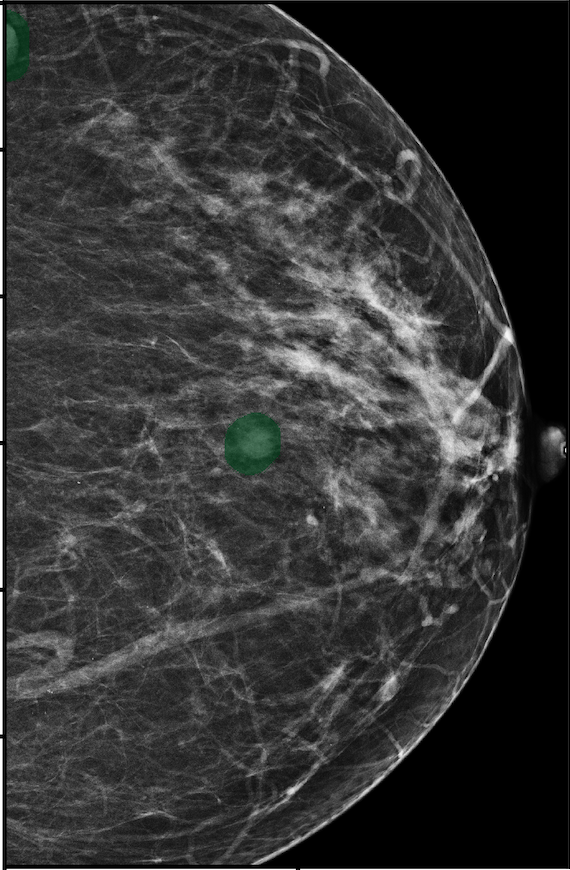}\vspace{-0.75mm}\\
    \footnotesize{R-CC} & \footnotesize{L-CC}\\
    \reflectbox{\includegraphics[width=0.14\textwidth]{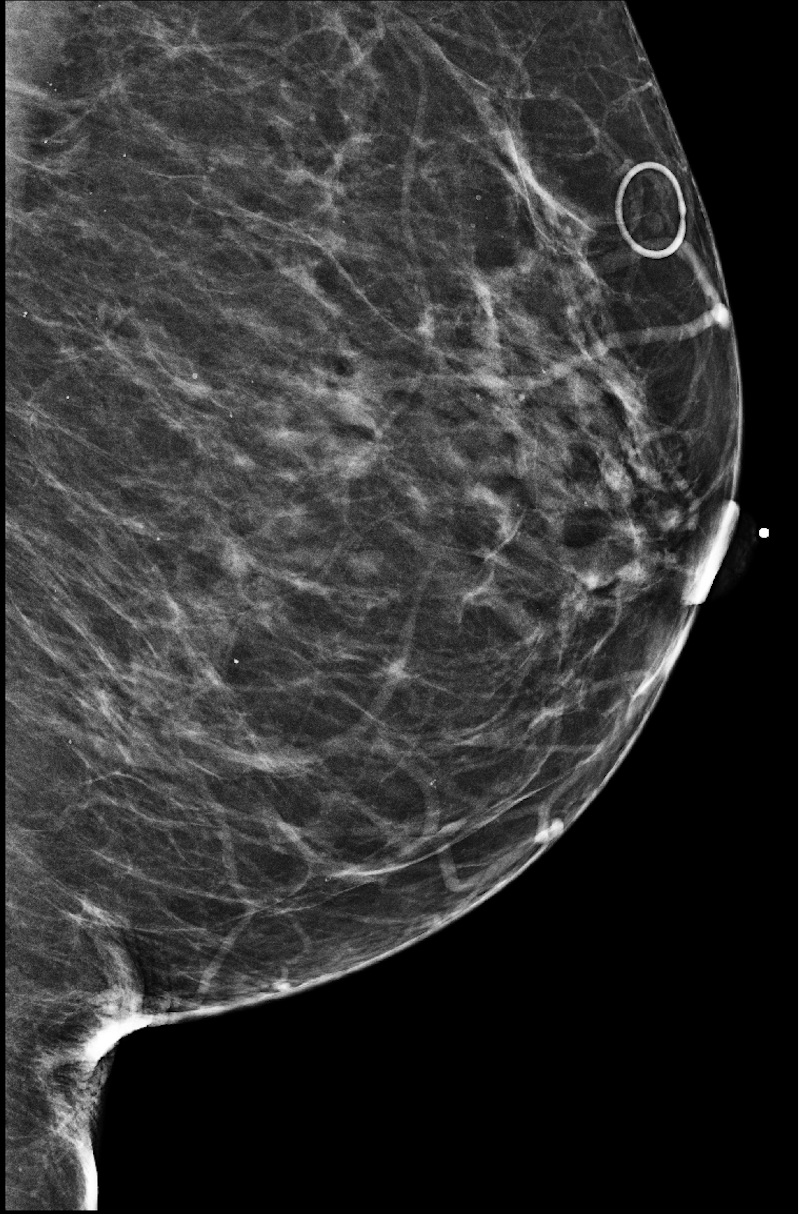}}\vspace{-0.75mm} &
    \includegraphics[width=0.14\textwidth]{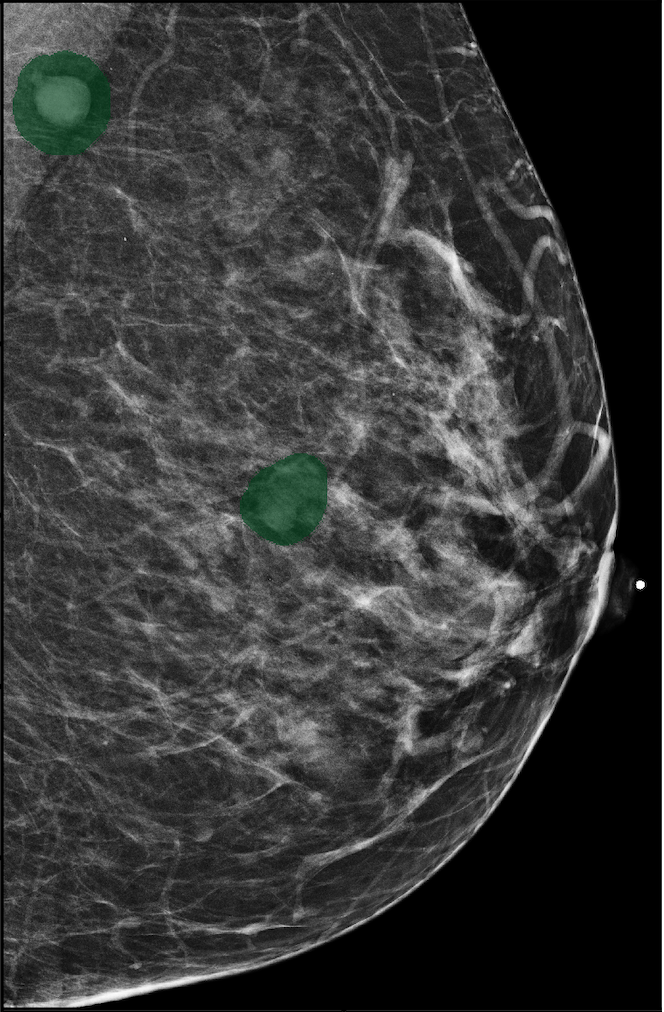}\vspace{-0.75mm} \\
    \footnotesize{R-MLO} & \footnotesize{L-MLO} \\
    \end{tabular}  
\vspace{-2.5mm}
\caption{Example exam for a patient. Benign findings are highlighted in green.}
\label{data_plot}
\vspace{-25pt}
\end{wrapfigure}

The proposed model is evaluated on the task of predicting whether any benign or malignant findings are present in a mammography exam. The dataset includes 229,426 exams (1,001,093 images). Across the entire data set, malignant findings were present in 985 breasts and benign findings in 5,556 breasts. As shown in Figure \ref{data_plot}, each exam contains four grayscale images ($2944 \times 1920$) representing two standard views (CC and MLO) for both left and right breasts. A label $\mathbf{y} \in \{0,1\}^2$ is associated with each breast where $y_c \in \{0,1\}$ ($c \in \{ \text{benign}, \text{malignant} \}$) denotes the presence or absence of a benign/malignant finding in a breast. All findings are confirmed by a biopsy. In each exam, two views on the same breast share the same label. A small fraction ($<1\%$) of the data are associated with pixel-level segmentation $\mathbf{M}^c \in \{0,1\}^{H \times W}$ where $\mathbf{M}^c_{i,j} = 1$ if pixel $i,j$ belongs to the findings of class $c$. In all experiments, segmentations are only used for evaluation.

\subsection{Experimental Set-up and Evaluation Metrics}
We adopt the same pre-processing as \cite{wu2019deep}. The dataset is divided into disjoint training (186,816), validation (28,462) and test (14,148) sets. In each iteration, we train the model using all exams that contain at least one benign or malignant finding and an equal number of randomly sampled negative exams. All images are cropped to $2944 \times 1920$ pixels and normalized. The training loss is optimized using Adam \cite{kingma2014adam}. We optimize the hyper-parameters using random search \cite{bergstra2012random}. Specifically, we search on a logarithmic scale for the learning rate $\eta \in 10^{[-5.5, -3.8]}$, the regularization weight $\lambda \in 10^{[-5, -2.8]}$, the regularization exponent $\beta \in e^{[-1.6, 1.6]}$, and the pooling threshold $t \in e^{[-5, -1.5]}$. We train 100 separate models, each for 40 epochs.

For classification performance, we report the area under the ROC curve (AUC) on the breast-level. As our model generates a prediction for each image and each breast is associated with two images (CC and MLO), we define breast-level predictions as the average of the two image-level predictions. To quantitatively evaluate our model's localization ability, we use the continuous F1 score, where precision (P) and recall (R) are defined as: $P = (\sum_{i,j \in \mathbf{M}^c} \mathbf{A}^c_{i,j})/(\sum_{i,j} \mathbf{A}^c_{i,j})$ and $R = (\sum_{i,j \in \mathbf{M}^c} \mathbf{A}^c_{i,j})/|\mathbf{M}^c|$, and $\mathbf{M}^c$ denotes the segmentation label and $\mathbf{A}^c$ is the SM for class $c$. On the test set, these metrics are averaged over images for which segmentation labels are available. 

\vspace{-10pt}
\begin{figure}
\parbox[t]{0.38\textwidth}{\null
  \centering
    \includegraphics[width=0.40\textwidth,keepaspectratio, trim=0 120 0 70]{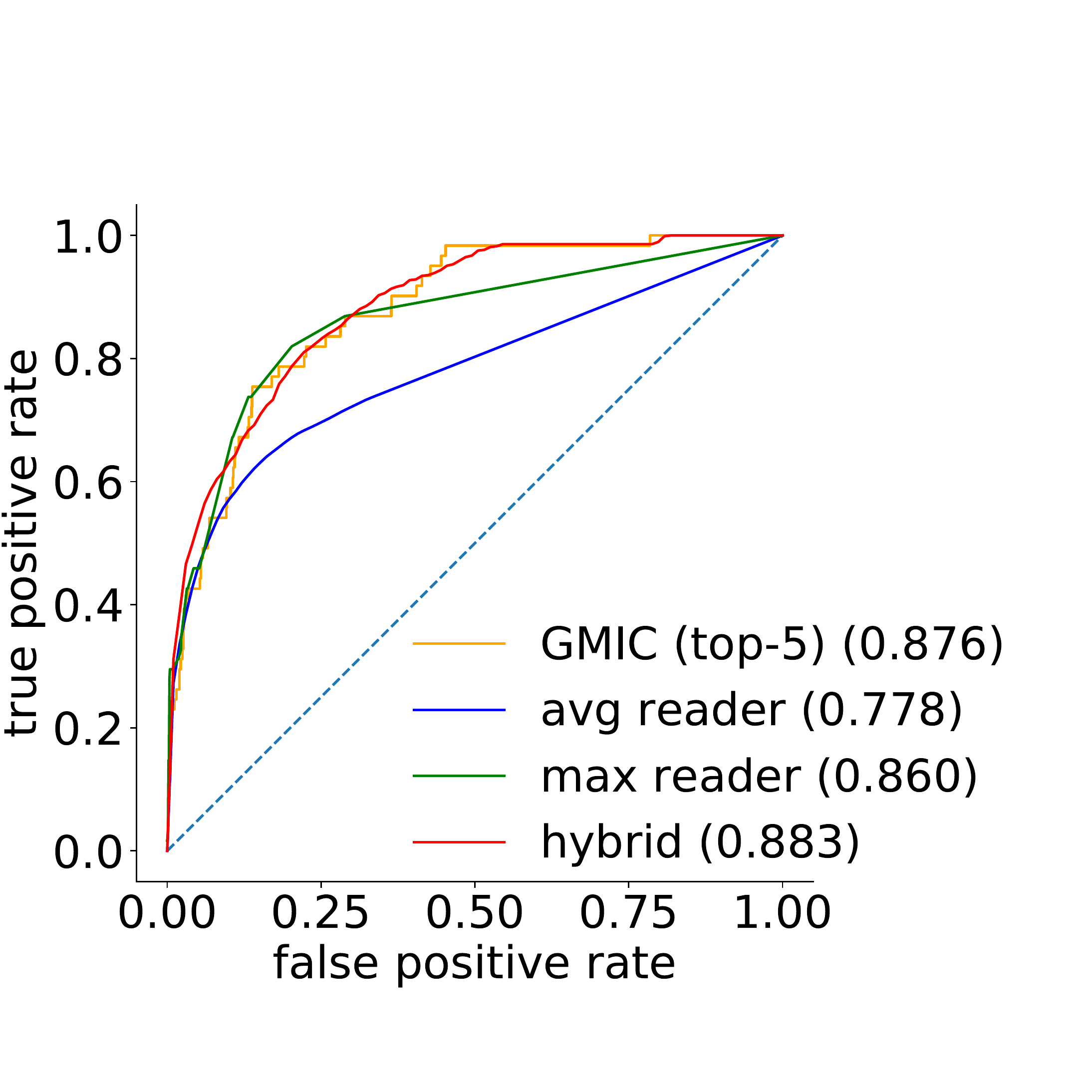}
    \label{fig:readerstudy}
    \captionof{figure}{Reader study}
}
\parbox[t]{0.6\textwidth}{\null
\centering
  \vskip-\abovecaptionskip
  \captionof{table}{AUCs of the baseline model and a few variations of GMIC}
  \vskip\abovecaptionskip
  \begin{tabular}{|l|c|c|}
    \hline
    Model & $ $ Malignant $ $ & $ $ Benign $ $  \\
    \hline
    ResNet-22 \cite{wu2019deep}  & 0.827 & 0.731 \\
    \Xhline{3\arrayrulewidth}
    GMIC-loc  & 0.885 & 0.777 \\
    \hline
    GMIC-mil  & 0.878 &  0.766  \\
    \hline
    GMIC-noattn  & 0.823  & 0.726  \\
    \hline
    GMIC-random  & 0.757  & 0.692  \\
    \hline
    GMIC-loc-random  & 0.889 & 0.776 \\
    \Xhline{3\arrayrulewidth}
    GMIC  &\textbf{0.900}  & \textbf{0.784} \\
    \hline
  \end{tabular}
  \label{table_AUC}
}
\end{figure}

\subsection{Classification Performance} \label{cp}
In this section, we report the average test performance of the 5 models from the hyper-parameter search that achieved the highest validation AUC on malignant classification (referred to as \textit{top-5}). In order to understand the impact of each module, we evaluate GMIC under a number of settings. GMIC-loc uses $\mathbf{\hat{y}}_\text{loc}$ as its predictions and GMIC-mil uses $\mathbf{\hat{y}}_\text{mil}$. As shown in Table 1, both variants of GMIC outperform the baseline, especially in predicting malignancy. The full model, GMIC, using the aggregated prediction $\mathbf{\hat{y}} = \frac{\mathbf{\hat{y}}_\text{loc}+\mathbf{\hat{y}}_\text{mil}}{2}$, attains higher AUC than GMIC-loc and GMIC-mil. We attribute this improvement to the synergy of local and global information. To empirically validate this conjecture, we test three additional models: GMIC-noattn assigns equal attentions on each ROI patch; GMIC-random outputs prediction $\mathbf{\hat{y}}_\text{random}$ by applying MIL module on patches randomly selected from the input image; GMIC-loc-random combines the predictions from GMIC-loc and GMIC-random $\mathbf{\hat{y}} = \frac{\mathbf{\hat{y}}_\text{loc} + \mathbf{\hat{y}}_\text{random}}{2}$. As Table 1 shows, GMIC-noattn is less accurate than GMIC-mil, suggesting that the attention mechanism in MIL module is essential for classification. Moreover, GMIC-random is weaker than GMIC-mil and GMIC-loc-random does not demonstrate any performance gain on top of GMIC-loc. These observations confirm our hypothesis that applying the MIL module on high-resolution ROI patches supplements the global information extracted by SM and refines predictions.

To evaluate the clinical value of our model, we compare the performance of GMIC with radiologists using data from the reader study described in \cite{wu2019deep}. This reader study includes 14 radiologists, each providing a probability estimate of malignancy for 720 screening exams (1440 breasts). The radiologists were only shown images for each exam with no other data. To further improve our predictions, we ensemble the predictions of the $\textit{top-5}$ models. As shown in Figure 3, the ensemble GMIC model achieves higher AUC (0.876) than the average (0.778) and the most accurate (0.860) among the 14 readers. GMIC obtains a marginally worse performance in the reader study than in the test set because the reader study contains a much larger portion of positive samples.

We also assess the efficacy of a human-machine hybrid, whose predictions are simply the average of predictions from the radiologists and the model. The human-machine hybrid achieves an AUC of 0.883. These results suggest that our model captures different aspects of the task compared to radiologists and can be used as a tool to assist in interpreting breast cancer screening exams. 

\begin{figure}[ht]
    \centering
\includegraphics[width=0.85\textwidth]{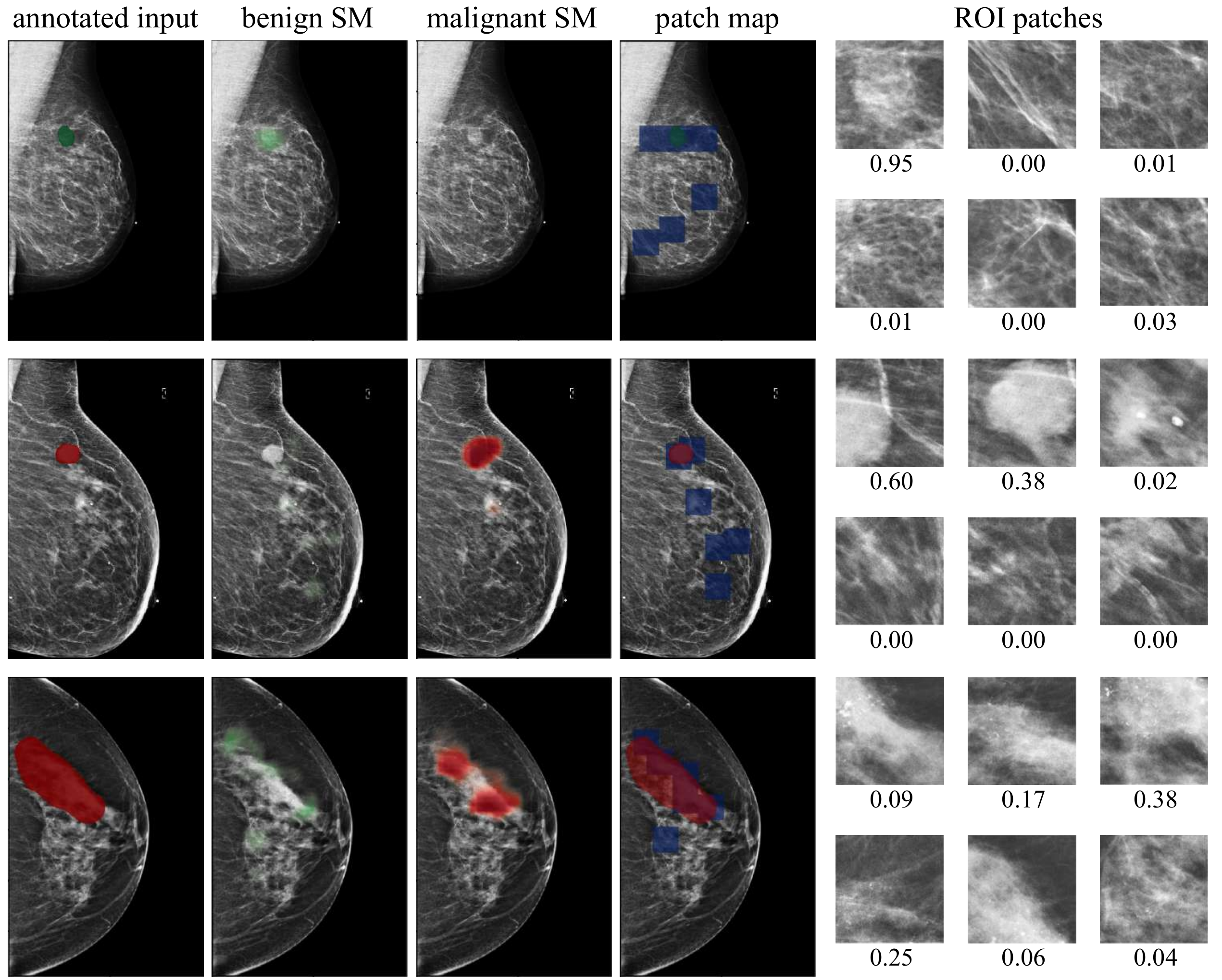}
    \caption{Visualization of three examples. Input images are annotated with segmentation labels (green=benign, red=malignant). ROI patches are shown with their attention scores.}
  \label{vis_plot}
  \vspace{-30pt}
\end{figure}

\subsection{Localization Performance}
We select the model with the highest validation F1 for malignancy localization. At the inference stage, we upsample SMs using nearest neighbour interpolation to match the resolution of the segmentation labels. The average continuous F1/precision/recall on test set is 0.207/0.288/0.254 for malignant and 0.133/0.135/0.224 for benign. In addition, the best localization model also achieves a classification AUC of 0.886/0.78 for malignant/benign classes. 

To better understand our model's behavior, we visualize SMs of three samples selected from the test set in Figure \ref{vis_plot}. In the first two examples, the SMs are highly activated on the true lesions, suggesting that our model is able to detect suspicious lesions without pixel-level supervision. Moreover, the attention $\alpha_k$ is highly concentrated on ROI patches that overlap with the annotated lesions. In the third example, the malignant SM only highlights parts of a large malignant lesion. This behavior is related to the design of $f_{\text{agg}}$: a fixed pooling threshold $t$ cannot be optimal for all sizes of ROI. Furthermore, this observation also illustrates that while human experts are asked to annotate the entire lesion, CNNs tend to emphasize only the most informative part. 

\section{Conclusion}
We present a novel model for breast cancer screening exam classification. The proposed method uses the input in its original resolution while being able to focus on fine details. Moreover, our model also generates saliency maps that provide additional interpretability. Evaluated on a large mammography dataset, GMIC outperforms the ResNet-based baseline and generates predictions that are as accurate as radiologists. Given its generic design, the proposed model is widely applicable to other image classification tasks. Our future research will focus on designing joint training mechanisms that would enable GMIC to improve its localization using error signals from the MIL module.\\
\vspace{-11pt}

\subsubsection*{Acknowledgments}
The authors would like to thank Catriona C. Geras for correcting earlier versions of this manuscript and Joe Katsnelson and Mario Videna for supporting our computing environment. We also gratefully acknowledge the support of Nvidia Corporation with the donation of some of the GPUs used in this research. This work was supported in part by grants from the National Institutes of Health (R21CA225175 and P41EB017183).


\clearpage
\appendix
\section{Additional Visualizations}
\begin{figure}
  \centering
 \includegraphics[width=0.66\textwidth, trim=0 0 0 20]{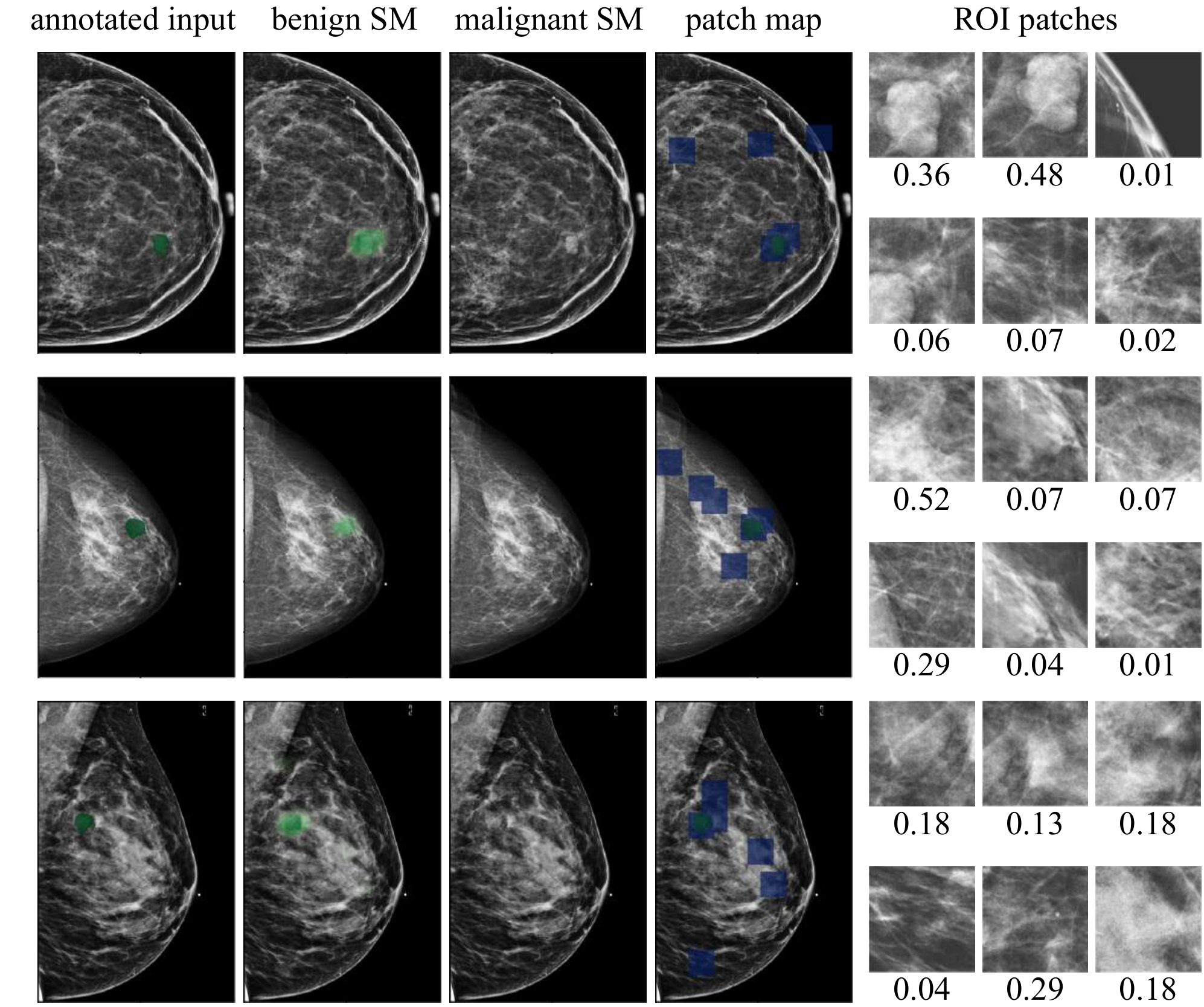}
 \hspace*{10pt}\includegraphics[width=0.64\textwidth, trim=0 0 0 -4]{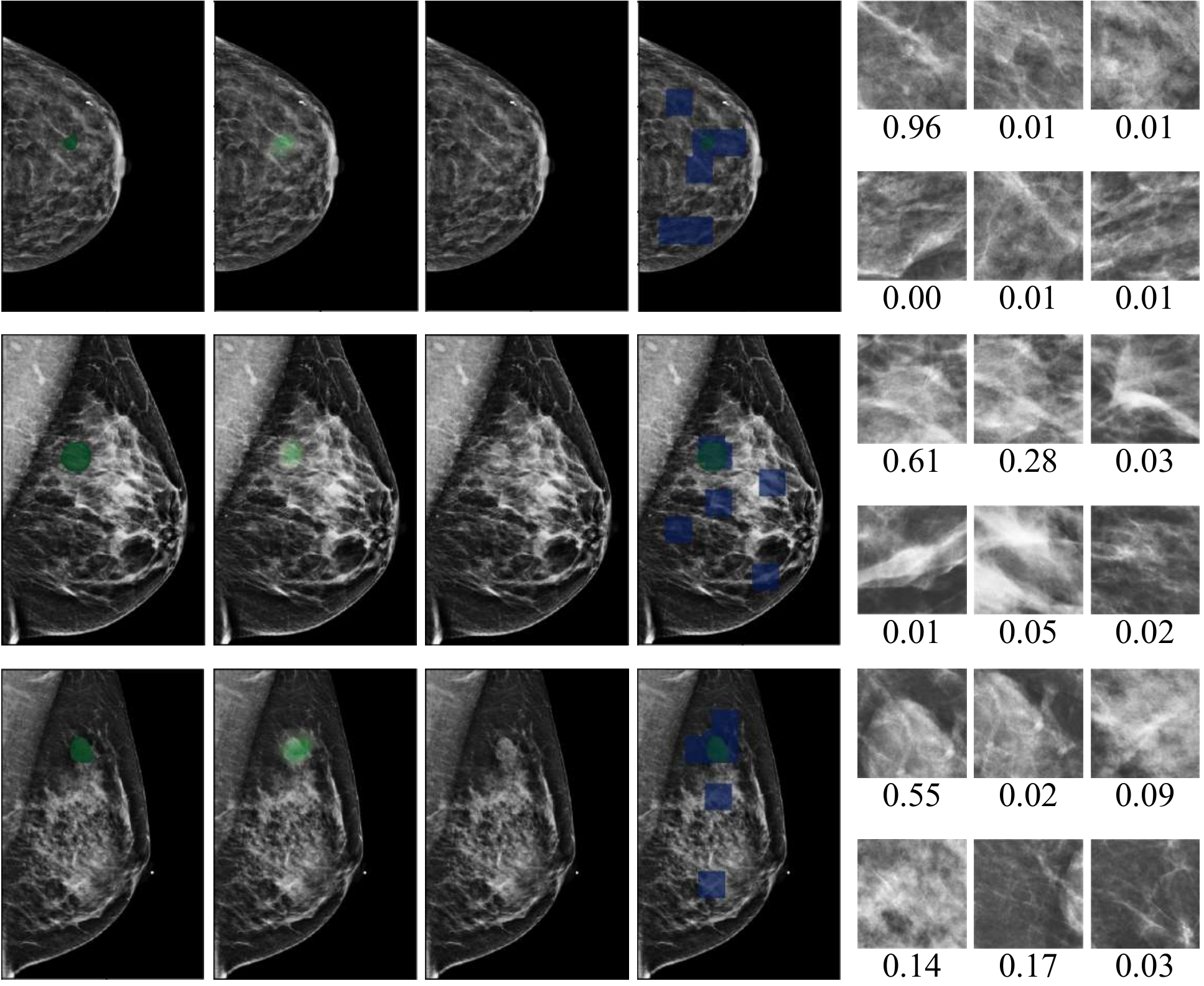}
  \caption{Additional visualizations of benign examples.  Input images are annotated with segmentation labels (green=benign, red=malignant). ROI patches are shown with their attention scores.}
\end{figure}

\begin{figure}
  \centering
 \includegraphics[width=0.66\textwidth, trim=0 0 0 20]{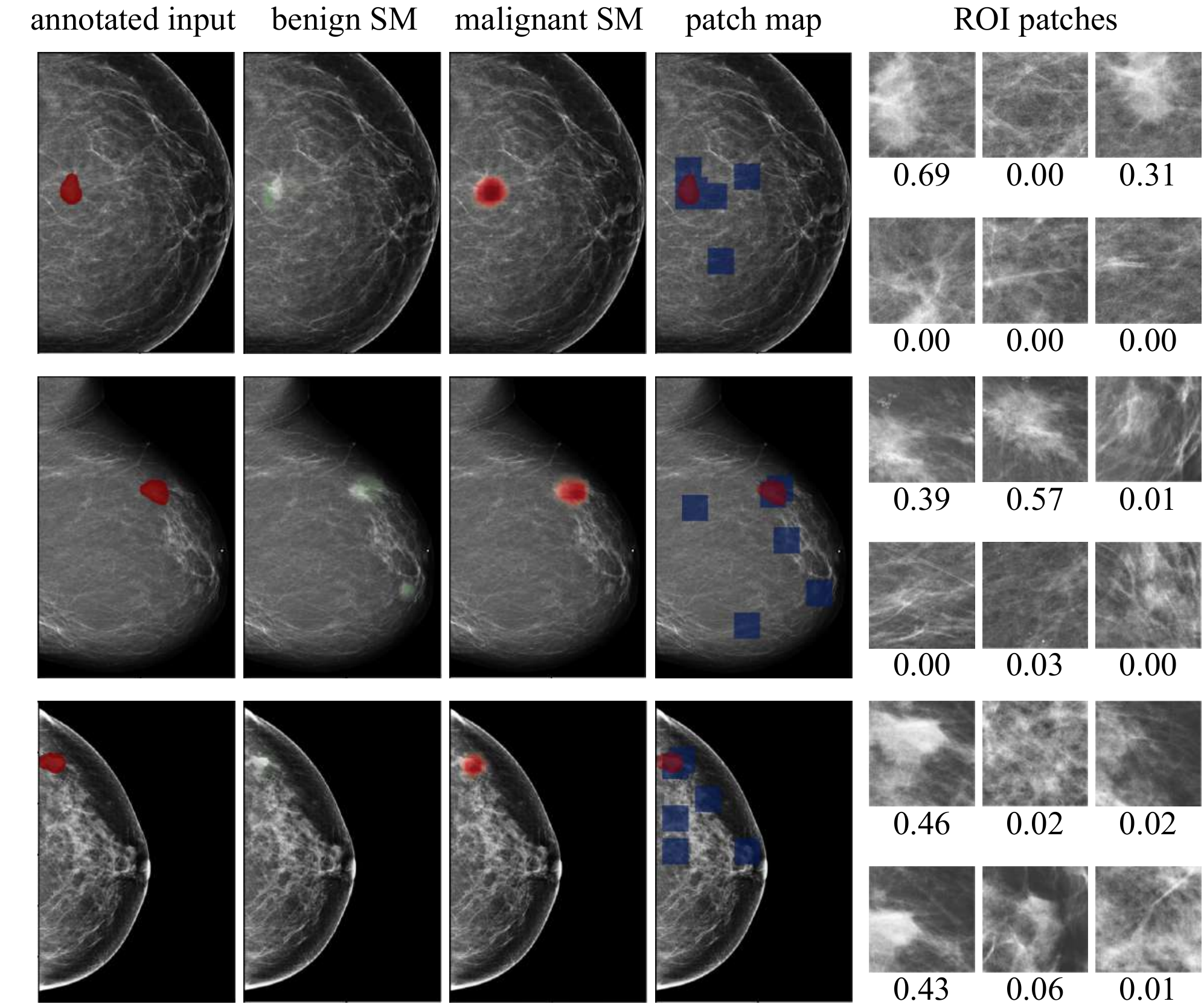}
 \hspace*{10pt}\includegraphics[width=0.64\textwidth, trim=0 0 0 -4]{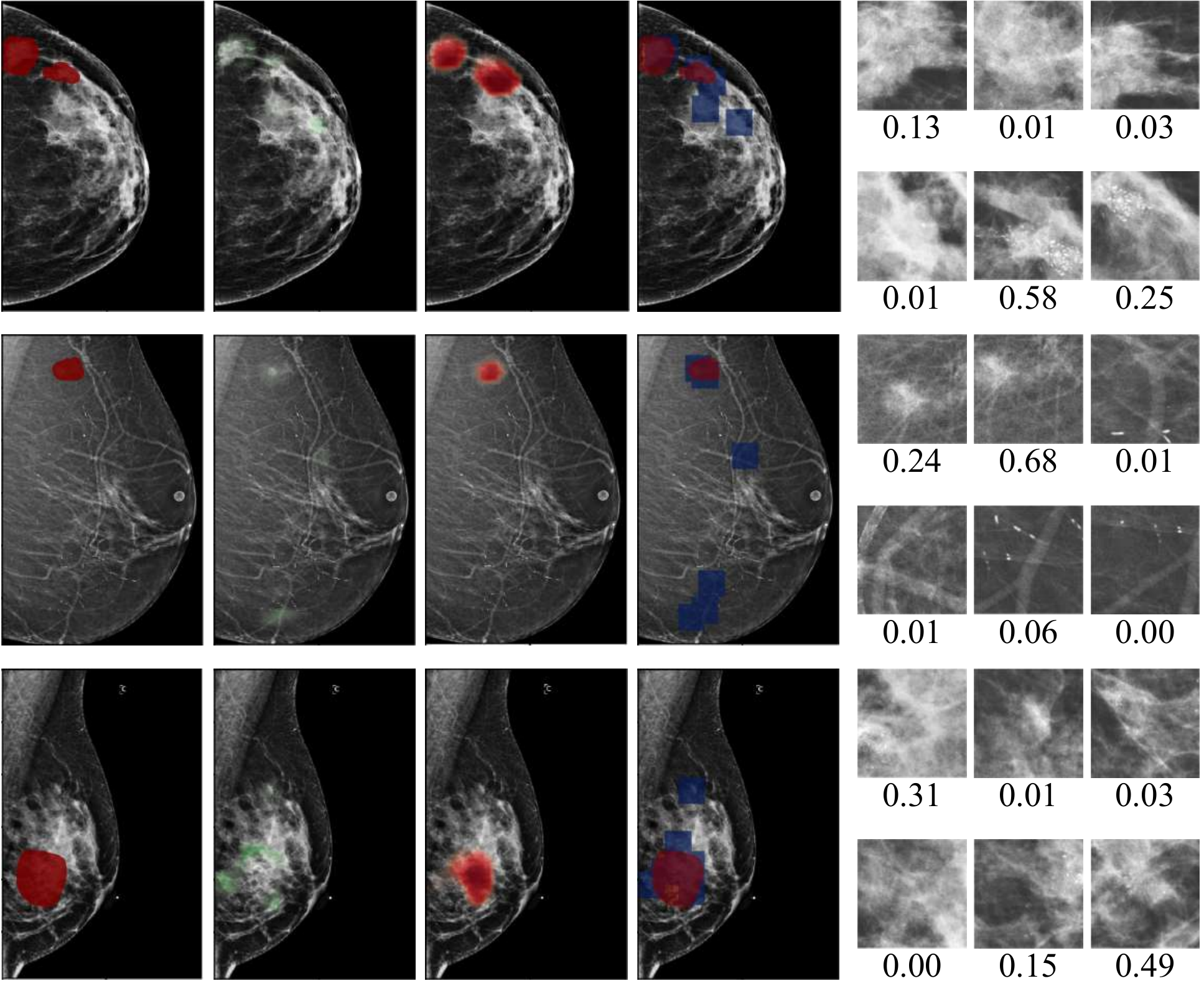}
  \caption{Additional visualizations of malignant examples.  Input images are annotated with segmentation labels (green=benign, red=malignant). ROI patches are shown with their attention scores.}
\end{figure}
\end{document}